\pdfoutput=1

\documentclass[11pt]{article}

\usepackage[]{ACL2023} 

\usepackage{times}
\usepackage{latexsym}
\usepackage{amsmath} 
\usepackage{booktabs}
\usepackage{multirow} 
\usepackage{graphicx}
\usepackage{array}
\usepackage{amsmath}  %
\usepackage{amssymb}  %

\newcolumntype{C}[1]{>{\centering\arraybackslash}p{#1}}

\usepackage[T1]{fontenc}

\usepackage[utf8]{inputenc}

\usepackage{microtype}

\usepackage{inconsolata}

\title{Aggregate, Decompose, and Fine-Tune: A Simple Yet Effective Factor-Tuning Method for Vision Transformer}

\author{
Dongping Chen \\ School of Computer Science and Technology\\ Huazhong University of Science and Technology\\ Wuhan, China\\
\texttt{u202112313@hust.edu.cn}
}

\begin{document}
\maketitle
\begin{abstract}
Recent advancements have illuminated the efficacy of some tensorization-decomposition Parameter-Efficient Fine-Tuning methods like $LoRA$ and $FacT$ in the context of Vision Transformers (ViT). However, these methods grapple with the challenges of inadequately addressing inner- and cross-layer redundancy. To tackle this issue, we introduce \textbf{EF}fective \textbf{F}actor-\textbf{T}uning (EFFT), a simple yet effective fine-tuning method. Within the VTAB-1K dataset, our $EFFT$  surpasses all baselines, attaining state-of-the-art performance with a categorical average of 75.9\% in top-1 accuracy with only 0.28\% of the parameters for full fine-tuning. Considering the simplicity and efficacy of $EFFT$, it holds the potential to serve as a foundational benchmark. The code and model are now available at \url{https://github.com/Dongping-Chen/EFFT-EFfective-Factor-Tuning}.
\end{abstract}

\begin{figure*}[htbp]
	\includegraphics[width=\linewidth]{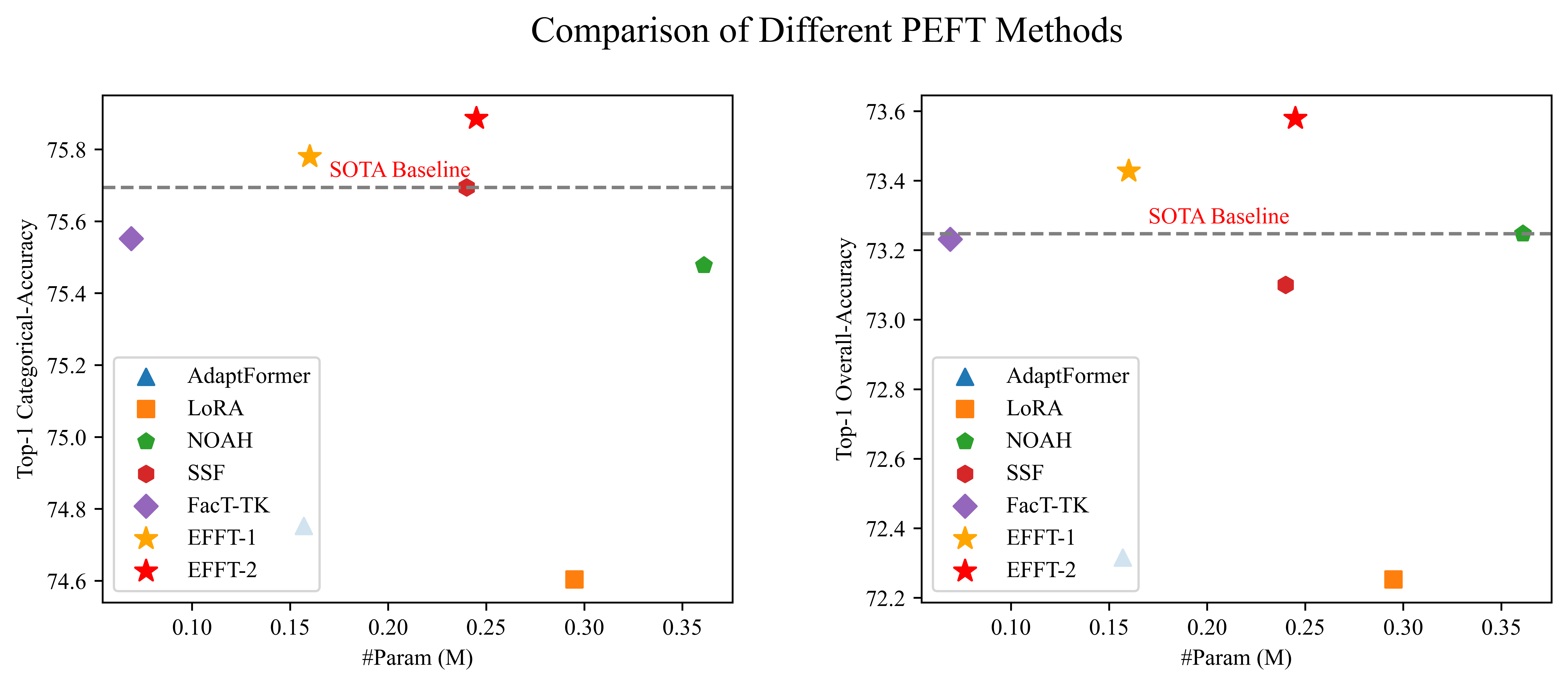}
	\caption{A comparative evaluation between our proposed $EFFT$ methodology and other mainstream PEFTs, namely $AdaptFormer$, $LoRA$, $NOAH$, $SSF$ and $FacT$, in terms of Top-1 categorical accuracy and overall accuracy on the VTAB-1K benchmark. Remarkably, $EFFT$ outperforms the SOTA baselines in both benchmarks, achieving this superior performance with the addition of only 0.16M and 0.25M extra parameters.}
	\label{fig:fig1}
\end{figure*}

\section{Introduction}
With the advent of architectures like the Transformer \citep{vaswani2017attention} and ResNet \citep{he2016deep}, there has been a significant escalation in model sizes, pushing the boundaries of computational feasibility. Several contemporary models, owing to their parameter magnitude, necessitate extended training durations, often spanning several months. This protraction poses challenges for the adaptation of pre-trained models to specialized downstream tasks, commonly referred to as fine-tuning. The computational intensity and protracted durations hinder the full fine-tuning of the model, advocating for parameter-efficient adaptations. Such methodologies entail selective parameter adjustments or the integration of minimal auxiliary networks \citep{ding2022delta,lialin2023scaling}. By adopting this approach, the majority of pre-trained parameters can remain fixed, necessitating modifications only to a limited set of task-specific parameters, thereby enhancing operational efficiency. 

Existing methodologies often incur time latency, primarily attributed to the incorporation of additional layers \citep{houlsby2019parameter,chen2022adaptformer,gui2023g,lester2021power}. Furthermore, certain approaches truncate the effective sequence length by reserving a portion exclusively for fine-tuning purposes \citep{jia2022visual}. Current tensorization-decomposition techniques \citep{hu2021lora,jie2023fact,chavan2023one,zhang2023adaptive} tend to neglect cross-layer associations, a notion elaborated upon by \citet{ren2022exploring}. This oversight results in parameter redundancy and inefficient storage, thereby impeding the effective distillation of core information. Current work \citep{kopiczko2023vera} utilizes vector-based random matrix adaptation for less trainable parameters, which is very suitable for large-scale models. Our research also aims to develop a redundancy-aware fine-tuning methodology that eliminates time latency and minimizes storage requirements for fine-tuning parameters.

Inspired by the work of \citet{jie2023fact}, we propose two novel and \textbf{EF}fective \textbf{F}actor-\textbf{T}uning (\textbf{EFFT}) methods. These methods are predicated on the principles of tensor decomposition, targeting the minimization of both inner- and cross-layer redundancies without introducing additional computational latency. By clustering matrices with akin attributes and consolidating them into a singular core tensor, our approach forges intrinsic connections within each group, enhancing the model's overall structure. Further, we conduct a comprehensive analysis of redundancy in vision transformers, exploring the subspace similarities among the decomposed matrices. This paper also includes ablation studies on variations across layers, blocks, and initialization techniques. Extensive experimentation across different Vision Transformer variants consistently confirms the superiority of the $EFFT$ method. The framework of our 
$EFFT$ method predominantly employs the Tensor-Train Format \citep{oseledets2011tensor}.

The primary contributions of our research are outlined as follows:
\begin{enumerate}
    \item We introduce the $EFFT$, designed as a plug-in for pre-trained models, facilitating adaptation to downstream tasks with its notable performance observed in transfer learning tasks.
    \item We evaluated $EFFT$ with vision transformers, and assessed its performance on the VTAB-1K benchmarks, achieving state-of-the-art (SOTA) results with 75.9\% categorical-average score in top-1 accuracy. See Figure \ref{fig:fig1} for a comparative evaluation.
\end{enumerate}

\section{Related Work}
\subsection{Parameter-Efficient Transfer Learning}
\noindent Contrary to full fine-tuning, parameter-efficient fine-tuning (PEFT) offers a way to adapt pre-trained models to specific downstream tasks efficiently. This approach has been gaining traction across various domains due to the significant savings in training time and memory. By training fewer parameters, PEFT methods can accommodate larger models or even allow for the utilization of bigger batch sizes. 

The idea of integrating $Adapter$ modules into transformer blocks was introduced by \citet{houlsby2019parameter}. These modules comprise two fully connected networks punctuated by a nonlinear activation. \citet{chen2022adaptformer} found that placing adapters just after MHSA yields performance akin to positioning them after both MHSA and FFN. However, these additional layers inevitably introduce excessive time latency for inference to the pre-trained model caused by the side path.

Soft prompts, as seen in methods like Prompt Tuning \citep{jia2022visual} and Prefix-Tuning \citep{li2021prefix}, offer another avenue that addresses embedding. While \citet{jia2022visual} augments input embedding with a trainable tensor, \citet{li2021prefix} introduces this tensor into the hidden states across all layers. Notably, in Prefix-Tuning, these soft prompts undergo processing by an FFN exclusively during training, ensuring the structural integrity of the pre-trained model is maintained. However, given the quadratic complexity of the transformer, these addictive methods significantly increase computation, shifting the downstream task-aware parameters to its additional space \citep{lialin2023scaling}.

\subsection{Tensorization-Decomposition Methods}

Tensor-decomposition techniques have emerged as effective strategies for achieving parameter efficiency in transformer-based models without time latency. 
$LoRA$ family \citep{hu2021lora,xu2023autolora,zhang2023adaptive,chavan2023one} stands as a notable approach. It deconstructs the additive parameter matrices in the transformer as $\Delta W = W_{down}W_{up}$. A significant advantage of $LoRA$ is its ability to compute in parallel with a pre-trained backbone, implying minimal latency overhead. 

Delving deeper, $FacT$ \citep{jie2023fact} employs both the Tensor-Train format and the Tucker format for decomposition tensor. This method achieves a remarkably reduced parameter count while simultaneously ensuring effective fine-tuning. The Tensor-Train variant of $FacT$, denoted as $FacT_{TT}$, can be expressed as:
\begin{equation}
\Delta W = s \cdot \Sigma \times U^{T} \times V^{T}
\end{equation}
where $\Sigma \in \mathbb{R}^{12L \times r_{1} \times r_{2}}$,\:$ U \in \mathbb{R}^{d \times r_{1}}$,\:$ V \in \mathbb{R}^{d \times r_{2}}$.

Although current tensorization-decomposition methods are robust and efficient, they still tend to neglect specific two kinds of redundancy within the model. While the $LoRA$ family primarily targets the decomposition and fine-tuning on the level of individual matrices, they fall short in shedding light on cross-layer redundancy that spans across multiple matrices of different layers. Although $FacT$ addresses the cross-layer problem, it offers too much freedom for each matrix and parameters for the core tensor and transcends the boundaries that separate MHSA and FFN blocks.
\begin{figure} 
	\includegraphics[width=\linewidth]{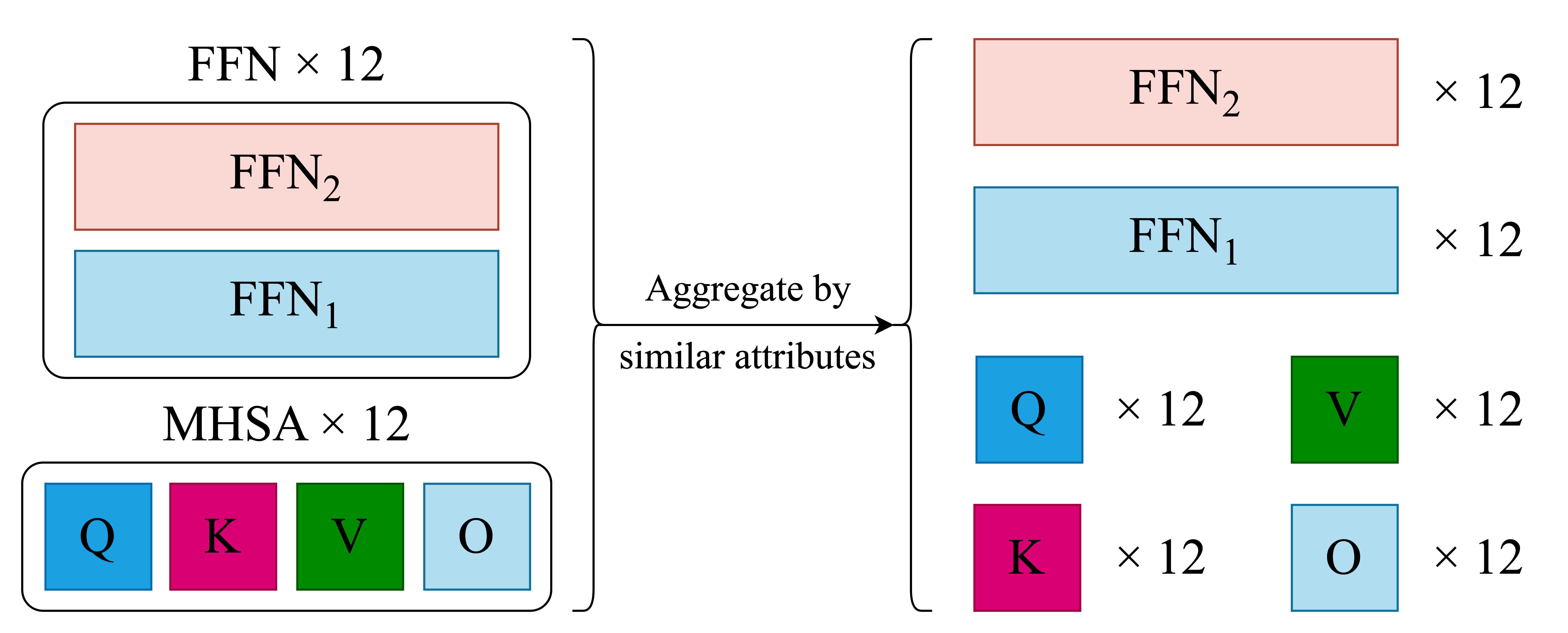}
	\caption{The figure elucidates the matrix aggregation process. As depicted, matrices with differing attributes and relative positions remain separate within the same block. Conversely, identical matrices from different layers are aggregated together.}
	\label{fig:fig2}
\end{figure}

\section{Methods}
\subsection{Aggregate the Similar Matrices}
Within large-scale models, identical structural elements can be observed. Given the inherent regularity of transformer architectures, it becomes viable to aggregate matrices among the transformer blocks. This aggregation facilitates subsequent decomposition and fine-tuning processes.

The transformer architecture, pivotal in modern NLP, comprises MHSA and FFN. Layer normalization and residual connections are integral and applied within each transformer block to maintain and propagate initial input information. Given an input $X \in \mathbb{R}^{n \times d}$, MHSA calculates attention scores using query ($Q$), key ($K$), and value ($V$) representations. Additionally, it employs projection matrices to aggregate multi-head outputs. These matrices are represented as $W_{q_{i}},\:W_{k_{i}},\:W_{v_{i}}  \in \mathbb{R}^{d\times h}$ and $W_{o} \in \mathbb{R}^ {d \times d}$. The attention computation can be expressed as follows:
\begin{gather}
Q = XW_{q_{i}}, 
K = XW_{k_{i}}, 
V = XW_{v_{i}}, \notag \\
\text{head}_{i} = \sum_{i=1}^{d} \text{Softmax} \left( \frac{Q \cdot K^{T}}{\sqrt{d_{h}}} \right) V, \notag \\
\text{MHSA}(X) = \text{Concat}(\text{head}_{1}, \dots, \text{head}_{n})W_{o} \notag
\end{gather}

Following MHSA, the output undergoes processing by FFN, which comprises two linear layers separated by a nonlinear activation. This structure aids in extracting prominent features from the attention mechanism. The process can be described as:
\begin{gather}
\text{FFN}(X) = \text{GeLU}(XW_{FFN_1})W_{FFN_2}, \notag \\
W_{FFN_1} \in \mathbb{R}^{d \times d_f}, 
W_{FFN_2} \in \mathbb{R}^{d_f \times d} \notag
\end{gather}
where $d_f=4d$ in $ViT$.

To eliminate cross-layer redundancy, we aggregate matrices from various parts of the model based on their similar attributes and relative positions. As illustrated in Figure \ref{fig:fig2}, We group $Q$,\:$K$,\:$V$, and $O$ matrices in MHSA, and two weight matrices in FFN across different layers. Drawing inspiration from $LoRA$, where a rank of 4 outperforms a rank of 64, we posit that constraining the freedom for each layer to distill core information can be advantageous. This leads us to train regularly structured matrices collectively, which is akin to equipping a plug-in, similar to $LoRA$, for an entire set of matrices on a group level that shares common attributes rather than applying it to a single matrix in isolation. 

\subsection{Decompose the Tensor}
Tensor decomposition is a technique that deconstructs a high-dimensional tensor into a series of lower-dimensional tensors. This methodology facilitates the consolidation of information, the extraction of fundamental features, and the reduction of the tensor's overall dimensions. Such objectives are congruent with the aims of PEFT, which concentrate on selectively updating a subset of parameters for particular downstream tasks.

\citet{allen2019can}, \citet{allen2020backward}, and~\citet{noach2020compressing} have demonstrated that weight matrices exhibit significant rank redundancy. This suggests that the core features pertinent to a specific task are considerably fewer than the entirety. \citep{hu2021lora} demonstrated that fine-tuning a minuscule subset of these core features can achieve state-of-the-art results, even though $LoRA$ predominantly focuses on the inner connections within a single matrix.

\begin{figure} [tbp]
	\includegraphics[width=\linewidth]{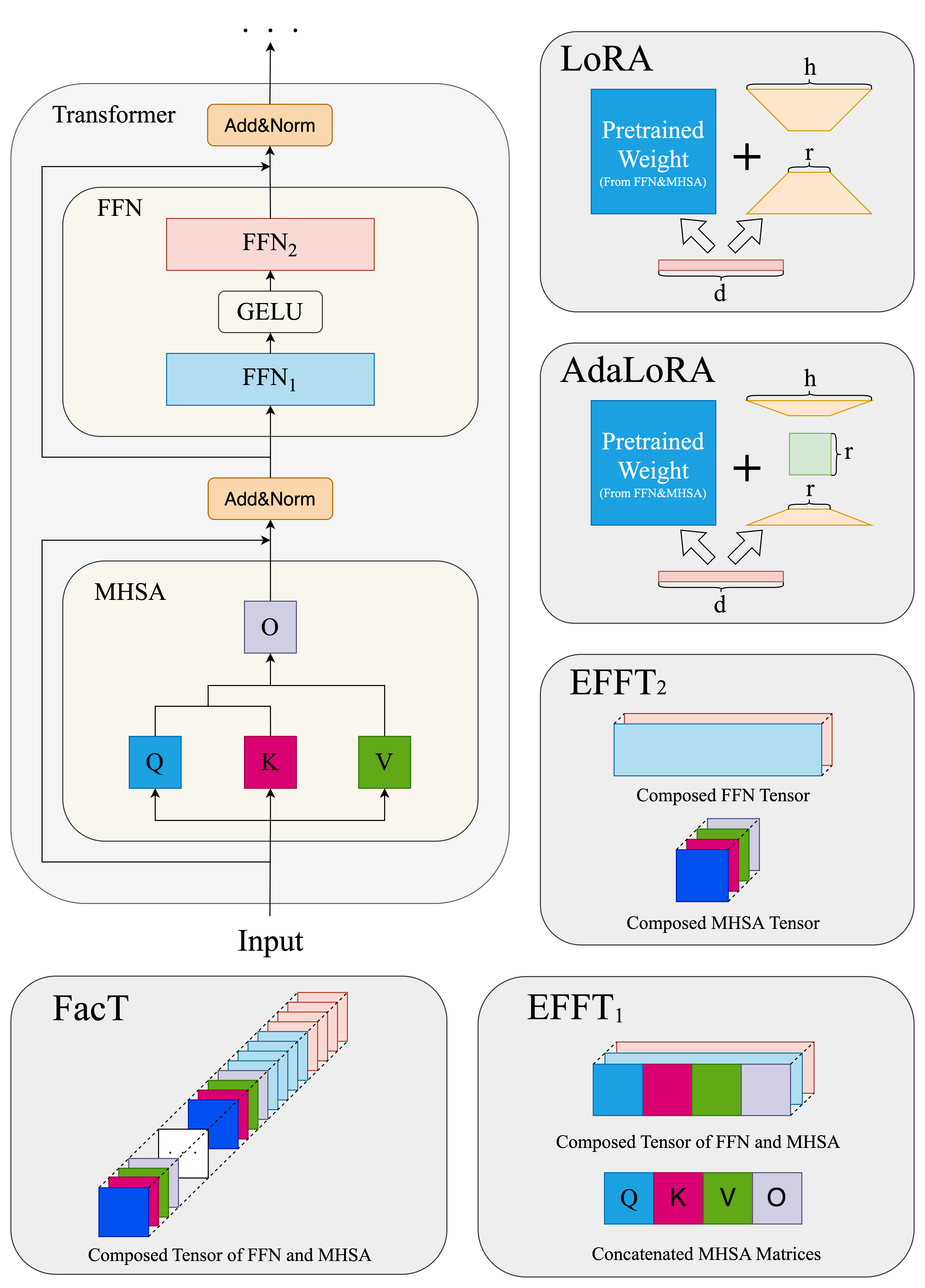}
	\caption{A comparative overview of tensorization-decomposition approaches. The figure's left section depicts a transformer block, which is comprised of MHSA and FFN. On the right side, we delineate the distinctions among the four PEFT methods.}
	\label{fig:fig3}
\end{figure}

As illustrated in Figure \ref{fig:fig3}, $EFFT_1$ concatenates four $d \times d$ matrices in MHSA into a $4d \times d$ matrix, which is then combined with $4d \times d$ matrices of FFN, resulting in a $3 \times 4d \times d$ tensor. It also emphasizes preserving the inner connections between MHSA and FFN. On the other hand, $EFFT_2$ condenses and decomposes MHSA and FFN separately, offering a more versatile approach by independently fine-tuning MHSA and FFN.

\begin{figure}[tbp] 
	\includegraphics[width=\linewidth]{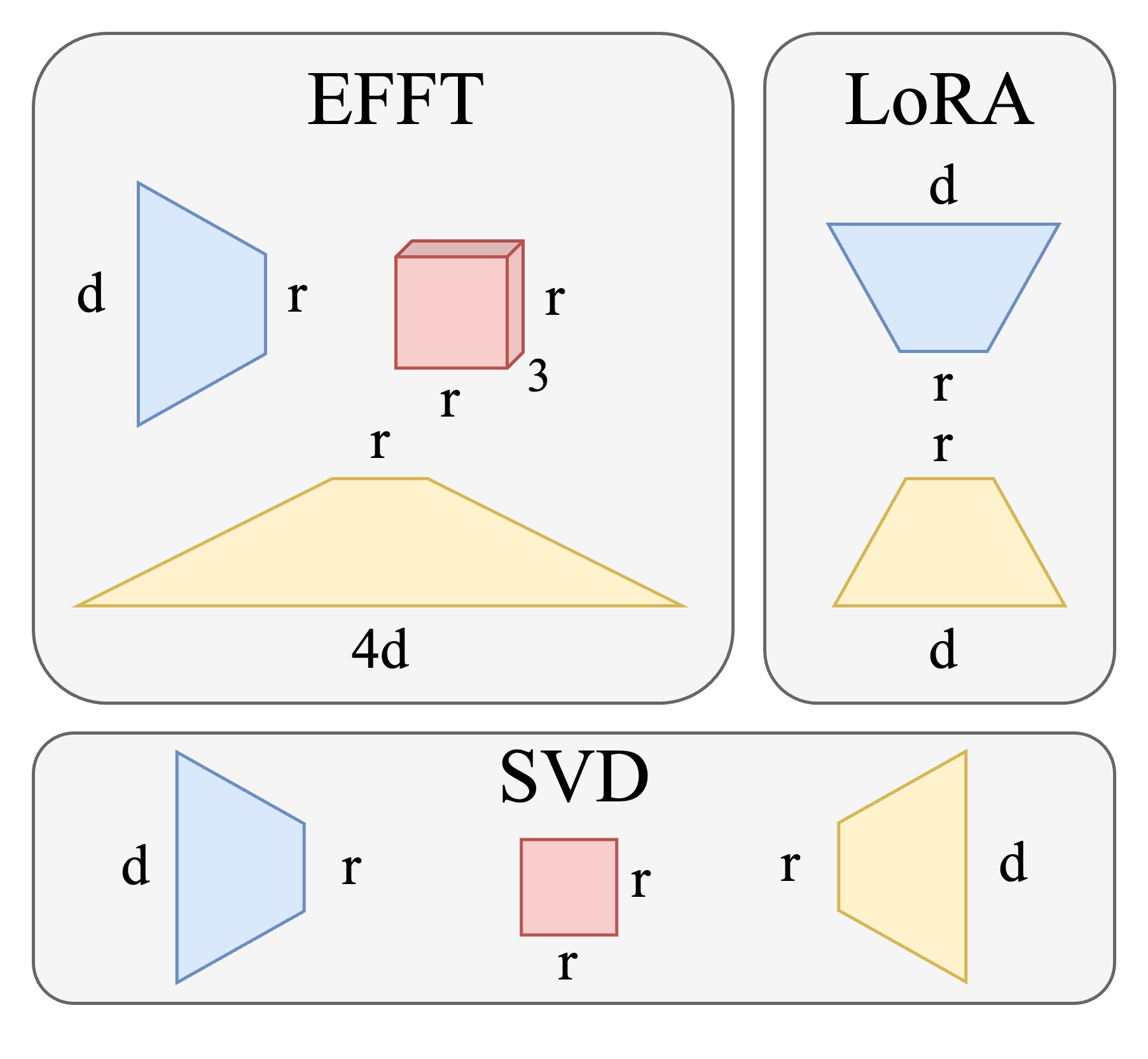}
	\caption{The figure compares feature extraction and redundancy elimination strategies across four tensorization-decomposition methods: $LoRA$, $SVD$, and $EFFT$. While $LoRA$ and $SVD$ focus on two-dimensional decomposition to capture core features along rows and columns, $EFFT$ extends to three dimensions.}
	\label{fig:fig4}
\end{figure}
Regarding the storage requirements for additional fine-tuning parameters, traditional methods typically fine-tune every parameter in the model, leading to an overhead of $O(Ld^2)$ parameters. However, by using the Tensor-Train Format, the storage of these additional parameters can be drastically reduced, as illustrated in Figure \ref{fig:fig4}.

\paragraph{Tensor-Train Format.}
In $EFFT_1$, where $\Delta W \in \mathbb{R}^{3 \times 4d \times d}$ can be decomposed as:
\begin{equation}
\Delta W = s \cdot \Sigma \times_{2} U^{T} \times_{3} V^{T}
\end{equation}
where $\Sigma \in \mathbb{R}^{3 \times r_1 \times r_2}, U \in \mathbb{R}^{4d \times r_1}, V \in \mathbb{R}^{d \times r_2}$ and $\times_i$ is a mode-$i$ product.

The tensor can be calculated as:
\begin{equation}
\Delta W_{i,j,k} = s \cdot \sum_{t_1=1}^{r_1} \sum_{t_2=1}^{r_2} \Sigma_{i,t_1,t_2} U_{j,t_1} V_{k,t_2}
\end{equation}
where $i \in \{1,2,3\}$,\:$j \in \{1,2, \dots, 4d\}$,\:$k \in \{1,2, \dots, d\}$,\:$ s \in 10^n$ and $n \in \mathbb{Z}$.

Given that $4d$ is 4 times larger than $d$, it suggests that more core features may reside in that dimension. This insight informs our strategy to set $r_1$ slightly larger to capture more information, leveraging more parameters to achieve higher accuracy. Following $FacT$'s pattern, we set $r=r_1=r_2<L << d$. The size of the additional parameters is approximately $5dr+3r^2 \sim O(dr+r^2)$, which is much smaller than full fine-tuning and marginally larger than both $FacT_{TT}$ and $FacT_{TK}$.

For $EFFT_2$, $\Delta W_{1} \in \mathbb{R}^{4 \times d \times d}$ and $\Delta W_{2} \in \mathbb{R}^{2 \times 4d \times d}$ are expressed as:
\begin{gather}
\Delta W_{1} = s_{1} \cdot \Sigma_{1} \times_{2} U_{1}^{T} \times_{3} V_{1}^{T},\notag \\
\Delta W_{2} = s_{2} \cdot \Sigma_{2} \times_{2} U_{2}^{T} \times_{3} V_{2}^{T} \notag
\end{gather}
where $\Sigma_{1} \in \mathbb{R}^{4 \times r_1 \times r_2}, U_{1} \in \mathbb{R}^{d \times r_1}, V_{1} \in \mathbb{R}^{d \times r_2}$ and $\Sigma_{2} \in \mathbb{R}^{2 \times r_1 \times r_2}, U_{2} \in \mathbb{R}^{4d \times r_1}, V_{2} \in \mathbb{R}^{d \times r_2}$.

In this scenario, the additional parameters amount to approximately $6dr+6r^2\sim O(dr+r^2)$, slightly larger than $EFFT_1$. Refer to Figure 5 for an analysis comparing feature extraction across the various methods.

\begin{figure} 
	\includegraphics[width=\linewidth]{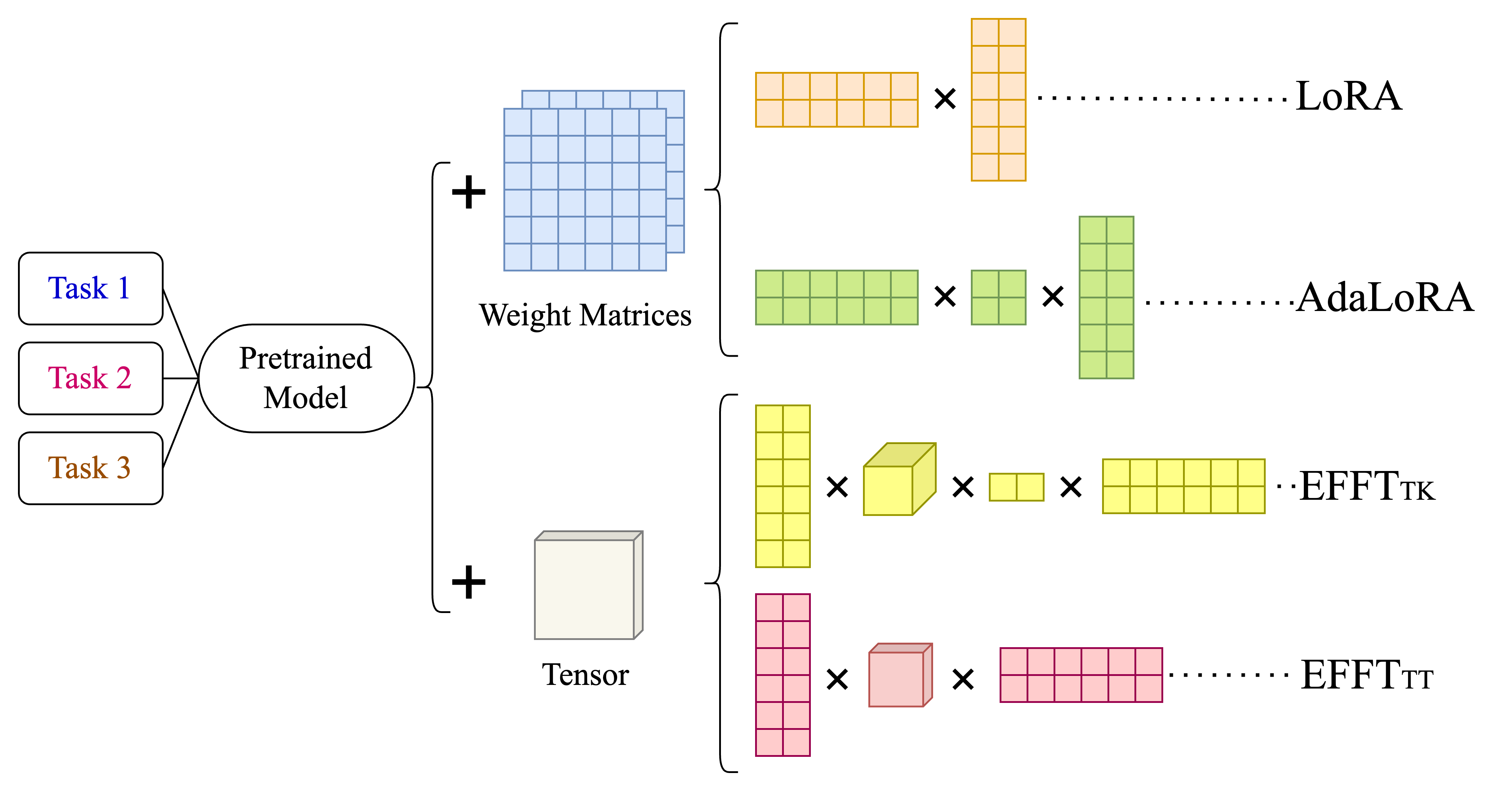}
	\caption{An In-Depth Visualization of the "Decompose-Then-Train" Paradigm, offering a detailed visual representation of the "decompose-then-train" approach, tailored for specific downstream tasks. $EFFT$ adopts a slightly different strategy: the parameters subject to training manifest as the same matrices and a core tensor.}
	\label{fig:fig5}
\end{figure}

\subsection{Fine-Tune the Model}
For $W(x) = W_0(x)+ \Delta W(x)$, the tensor $\Delta W$ is constructed from several smaller tensors that are updated during the training phase. Our work follows the "decompose-then-train" paradigm, as illustrated in Figure \ref{fig:fig5}, which trains the low-rank adaptation matrices and pivotal tensor together. By aggregating matrices with analogous characteristics, $EFFT$ effectively mitigates redundancy from inner- and cross-layer connections. This implies that, unlike the $LoRA$  where additional trainable parameters are matrix-specific, our method introduces cross-layer parameters.

Among our two decomposition techniques, we initialize only one matrix to zero, while the remaining matrices are subject to random normalization. This ensures that $\Delta W$ starts as a zero tensor, implying no initial impact on the model. We subsequently find that initializing either $U$ or $V$ to zero does have an impact on the learning capability of our model. Interestingly, $V$ appears to have a marginally superior effect compared to $U$.

\subsection{Analysis Method of Subspace Similarities}

To investigate the rationality of $EFFT$, which concatenates MHSA and FFN together, this paper conducted several experiments focusing on the subspace similarities. Given that the additional tensor is a product of decomposed tensors, we can infer the overall tensor similarity by calculating the similarities of these small matrices.

 Following the methodology utilized in \citet{hu2021lora}, we seek to quantify the similarity of the subspace spanned by the top-$i$ left singular vectors in various configurations. We employ a normalized subspace similarity metric based on Grassmann Distance, which generally applies to vectors of different dimensions. Let $U_A^i$ denote the top-$i$ left singular vectors of matrix A. The similarity is then calculated as:
\begin{equation}
\text{Similarity}(A, B, i, j) = \frac{\| U_i^T U_j \|_F^2}{\min(i,j)} \in [0,1]
\end{equation}

\section{Experiments}
\begin{figure*}[h] 
\centering
	\includegraphics[width=\linewidth]{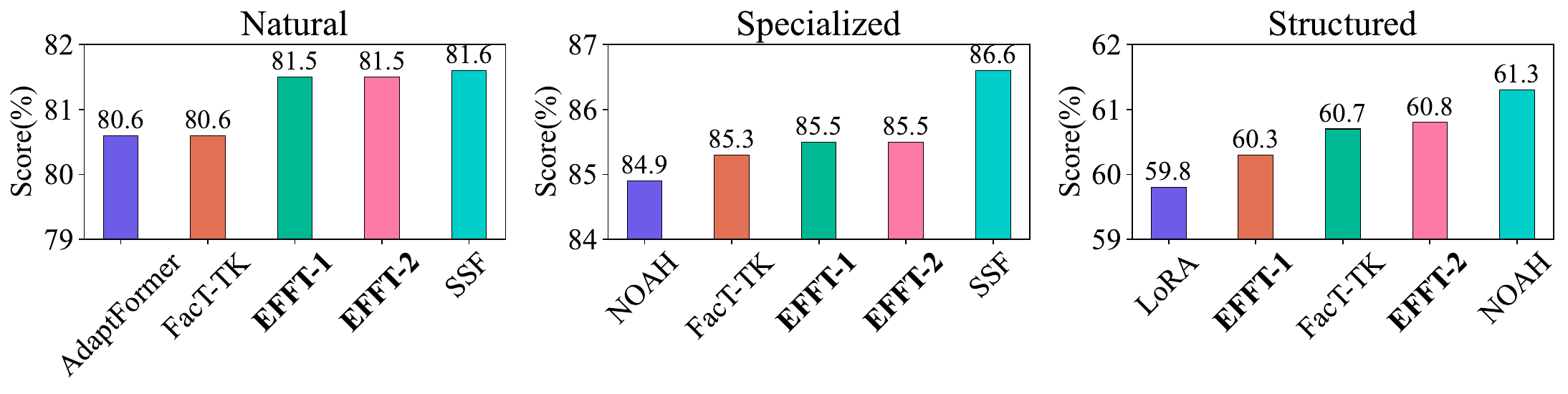}
	\caption{A comparative analysis of PEFTs on categorical-average: $EFFT$ surpasses other PEFTs in the Natural.}
	\label{fig:fig6}
\end{figure*}

For our experimental framework, we initially subject our $EFFT$  to rigorous evaluation using the VTAB-1K benchmark, to evaluate its performance against multiple SOTA baseline models. Subsequently, we extend our evaluation to encompass different vision transformers \citep{dosovitskiy2020image} and hierarchical swin transformers\citep{liu2021swin}. In the concluding phase of our experimental setup, we undertake ablation studies and delve into the intricacies of subspace similarities within the "decompose-then-train" approach. As shown in Table \ref{tab:table1} and Figure \ref{fig:fig6}, our $EFFT$ outperforms other PEFT methods and finally reaches some interesting conclusions.

\begin{table*}[t]
\fontsize{8pt}{12pt}\selectfont
\begin{tabular}{C{0.9cm}C{0.5cm}|C{0.2cm}C{0.2cm}C{0.2cm}C{0.2cm}C{0.2cm}C{0.2cm}C{0.4cm}|C{0.2cm}C{0.2cm}C{0.2cm}C{0.4cm}|C{0.2cm}C{0.2cm}C{0.2cm}C{0.2cm}C{0.2cm}C{0.2cm}C{0.2cm}C{0.4cm}|C{0.2cm}C{0.2cm}}
\toprule[2pt]
\multirow{2}{*}{\rotatebox{90}{{Name~~~~~~~~~~~~}}} & \multirow{2}{*}{\rotatebox{90}{\#Param (M)~~~~~~~~}} & \multicolumn{7}{c|}{\textbf{Natural}}                                                                                  & \multicolumn{4}{c|}{\textbf{Specialized}}                               & \multicolumn{8}{c|}{\textbf{Structured}}                                                                                        & \multicolumn{2}{c}{Average}     \\
                      &                              & \rotatebox{90}{cifar}         & \rotatebox{90}{caltech101}    & \rotatebox{90}{dtd}           & \rotatebox{90}{flowers102} & \rotatebox{90}{iiit\_pet} & \rotatebox{90}{svhn}          & \rotatebox{90}{sun397}        & \rotatebox{90}{patch\_camelyon} & \rotatebox{90}{eurosat}       & \rotatebox{90}{resisc45}      & \rotatebox{90}{diabetic} & \rotatebox{90}{clevr\_count}  & \rotatebox{90}{clevr\_dist}   & \rotatebox{90}{dmlab}         & \rotatebox{90}{kitti}         & \rotatebox{90}{dsprites\_loc} &
                      \rotatebox{90}{dsprites\_ori} & \rotatebox{90}{smallnorb\_azi} & \rotatebox{90}{smallnorb\_ele} & \rotatebox{90}{Overall Ave.}   & \rotatebox{90}{Subsets Ave.} \\ \midrule[1.5pt]
Full                  & 85.8                         & 68.9          & 87.7          & 64.3          & 97.2               & 86.9              & 87.4          & 38.8          & 79.7            & 95.7          & 84.2          & 73.9                  & 56.3          & 58.6          & 41.7          & 65.5          & 57.5          & 46.7          & 25.7           & 29.1           & 65.6          & 69.0            \\
Linear        & 0                            & 64.4          & 85.0          & 63.2          & 97.0               & 86.3              & 36.6          & 51.0          & 78.5            & 87.5          & 68.5          & 74.0                  & 34.3          & 30.6          & 33.2          & 55.4          & 12.5          & 20.0          & 9.6            & 19.2           & 53.0          & 57.7            \\ \midrule[1.5pt]
$BitFit$                & 0.103                        & 72.8          & 87.0          & 59.2          & 97.5               & 85.3              & 59.9          & 51.4          & 78.7            & 91.6          & 72.9          & 69.8                  & 61.5          & 55.6          & 32.4          & 55.9          & 66.6          & 40.0          & 15.7           & 25.1           & 62.0          & 65.2            \\
$VPT_{S}$           & 0.063                        & \textbf{77.7} & 86.9          & 62.6          & 97.5               & 87.3              & 74.5          & 51.2          & 78.2            & 92.0          & 75.6          & 72.9                  & 50.5          & 58.6          & 40.5          & 67.1          & 68.7          & 36.1          & 20.2           & 34.1           & 64.9          & 67.8            \\
$VPT_{D}$              & 0.531                        & \textbf{78.8} & 90.8          & 65.8          & 98.0               & 88.3              & 78.1          & 49.6          & 81.8            & \textbf{96.1} & 83.4          & 68.4                  & 68.5          & 60.0          & 46.5          & 72.8          & 73.6          & 47.9          & 32.9           & 37.8           & 69.4          & 72.0            \\
$Adapter$               & 0.157                        & 69.2          & 90.1          & 68.0          & 98.8               & 89.9              & 82.8          & 54.3          & 84.0            & 94.9          & 81.9          & 75.5                  & 80.9          & 65.3          & 48.6          & 78.3          & 74.8          & 48.5          & 29.9           & 41.6           & 71.4          & 73.9            \\
$AdaptF$           & 0.157                        & 70.8          & 91.2          & 70.5          & \textbf{99.1}      & \textbf{90.9}     & 86.6          & \textbf{54.8} & 83.0            & 95.8          & 84.4          & \textbf{76.3}         & 81.9          & 64.3          & 49.3          & 80.3          & 76.3          & 45.7          & 31.7           & 41.1           & 72.3          & 74.8            \\
$LoRA$                  & 0.295                        & 67.1          & 91.4          & 69.4          & 98.8               & 90.4              & 85.3          & 54.0          & 84.9            & 95.3          & 84.4          & 73.6                  & \textbf{82.9} & \textbf{69.2} & 49.8          & 78.5          & 75.7          & 47.1          & 31.0           & \textbf{44.0}  & 72.3          & 74.6            \\
$NOAH$                  & 0.361                        & 69.6          & 92.7          & 70.2          & \textbf{99.1}      & 90.4              & 86.1          & 53.7          & 84.4            & 95.4          & 83.9          & 75.8                  & \textbf{82.8} & \textbf{68.9} & 49.9          & \textbf{81.7} & \textbf{81.8} & 48.3          & 32.8           & \textbf{44.2}  & \textbf{73.2} & 75.5            \\
$SSF$                   & 0.24                         & 69.0          & 92.6          & \textbf{75.1} & \textbf{99.4}      & \textbf{91.8}     & \textbf{90.2} & 52.9          & \textbf{87.4}   & 95.9          & \textbf{87.4} & 75.5                  & 75.9          & 62.3          & \textbf{53.3} & 80.6          & 77.3          & \textbf{54.9} & 29.5           & 37.9           & 73.1          & \textbf{75.7}   \\
$FacT^{TK}_{\leq 32}$               & 0.069                        & 70.6          & 90.6          & 70.8          & \textbf{99.1}      & 90.7              & 88.6          & 54.1          & 84.8            & \textbf{96.2} & \textbf{84.5} & 75.7                  & \textbf{82.6} & \textbf{68.2} & 49.8          & 80.7          & \textbf{80.8} & 47.4          & \textbf{33.2}  & \textbf{43.0}  & \textbf{73.2} & 75.6            \\ \midrule[1.5pt]
\textbf{$EFFT^1_{16}$}    & 0.076                        & 71.5          & 92.5          & 70.2          & 99.0               & 90.6              & 88.5          & 54.2          & 84.1            & \textbf{96.1} & 84.3          & \textbf{75.9}         & 78.4          & 66.1          & \textbf{50.1} & 81.2          & 79.1          & \textbf{48.6} & 32.2           & 41.4           & 72.8          & 75.2            \\
\textbf{$EFFT^2_{16}$}    & 0.144                        & 72.1          & \textbf{92.8} & 70.7          & \textbf{99.1}      & 90.6              & 87.5          & \textbf{54.8} & 85.2            & 95.7          & 84.4          & \textbf{75.9}         & 80.3          & 66.2          & \textbf{50.1} & \textbf{82.7} & 78.3          & 47.7          & 32.7           & 41.0           & 73.0          & 75.4            \\
\textbf{$EFFT^1_{\leq 32}$}    & 0.16                         & 72.4          & \textbf{93.1} & \textbf{71.1} & \textbf{99.1}      & \textbf{90.8}     & \textbf{89.6} & 54.4          & \textbf{85.5}   & \textbf{96.1} & \textbf{84.6} & \textbf{75.9}         & 80.0          & 66.3          & \textbf{50.1} & 81.4          & \textbf{79.4} & \textbf{48.6} & \textbf{35.0}  & 41.9           & \textbf{73.4} & \textbf{75.8}   \\
\textbf{$EFFT^2_{\leq 32}$}    & 0.245                        & 72.1          & \textbf{92.9} & \textbf{71.2} & \textbf{99.1}      & 90.6              & \textbf{89.6} & \textbf{54.7} & \textbf{85.4}   & 95.7          & 84.4          & \textbf{76.3}         & 81.6          & 67.6          & \textbf{50.4} & \textbf{82.7} & 79.2          & 47.7          & \textbf{34.9}  & 41.9           & \textbf{73.6} & \textbf{75.9}   \\ \bottomrule[2pt]
\end{tabular}
\caption{Full results on the VTAB-1K benchmark. $\#params$ specifies the number of trainable parameters in backbones.  Average accuracy is averaged over group-wise average values. Our $EFFT-TK_{\leq 32}$ outperforms all previous methods while using fewer parameters.}
\label{tab:table1}
\end{table*}

\subsection{Fine-Tuning on VTAB-1K benchmark}
For all conducted experiments, we adopt a ViT-B/16 model pre-trained on the supervised ImageNet-21k \citep{deng2009imagenet} as the backbone architecture, which produces an excellent foundation for transfer learning tasks. We leverage the VTAB-1K benchmark to evaluate the effectiveness of our methods. VTAB-1K comprises 19 distinct visual classification datasets, categorized into Natural, Specialized, and Structured subsets. Each dataset is restricted to 1,000 training samples.

\paragraph{Implementation Details.}
We compare our $EFFT$ against a variety of robust baselines, including $BitFit$ \citep{zaken2021bitfit}, $VPT_{shallow}$ \& $VPT_{deep}$ \citep{jia2022visual}, $AdapterFormer$ \citep{chen2022adaptformer}, $LoRA$ \citep{hu2021lora}, $NOAH$ \citep{zhang2022neural}, $FacT$ \citep{jie2023fact} and the current SOTA method, $SSF$ \citep{lian2022scaling}. Following guidelines set forth by \citet{zhang2022neural} and \citet{jie2023fact}, we set the hidden dimension $h$ for both $Adapter$ and $AdapterFormer$, as well as $r$ for $LoRA$, to a value of 8. For $VPT$, we adhere to the prompt length $l$ as specified in its original publication. For $FacT$, we propose its best performance $FacT_{\leq 32}^{TK}$. Additionally, we include two conventional transfer learning baselines: Full fine-tuning and linear probing, which involves training a linear classification layer atop the pre-trained model.

For our methods, we have four distinct configurations:
$EFFT_1$ and $EFFT_2$ with $r$ = 16; $EFFT^1_{\leq 32}$ and $EFFT^1_{\leq 32}$ where $r$ is selected from $\{8,16,32\}$. We employ the AdamW optimizer with a learning rate of 1e-3 and a batch size of 64 for 100 epochs. The hyper-parameter $s$ is empirically chosen from $\{0.1, 1, 10, 100\}$.

\begin{figure*} 
	\includegraphics[width=\linewidth]{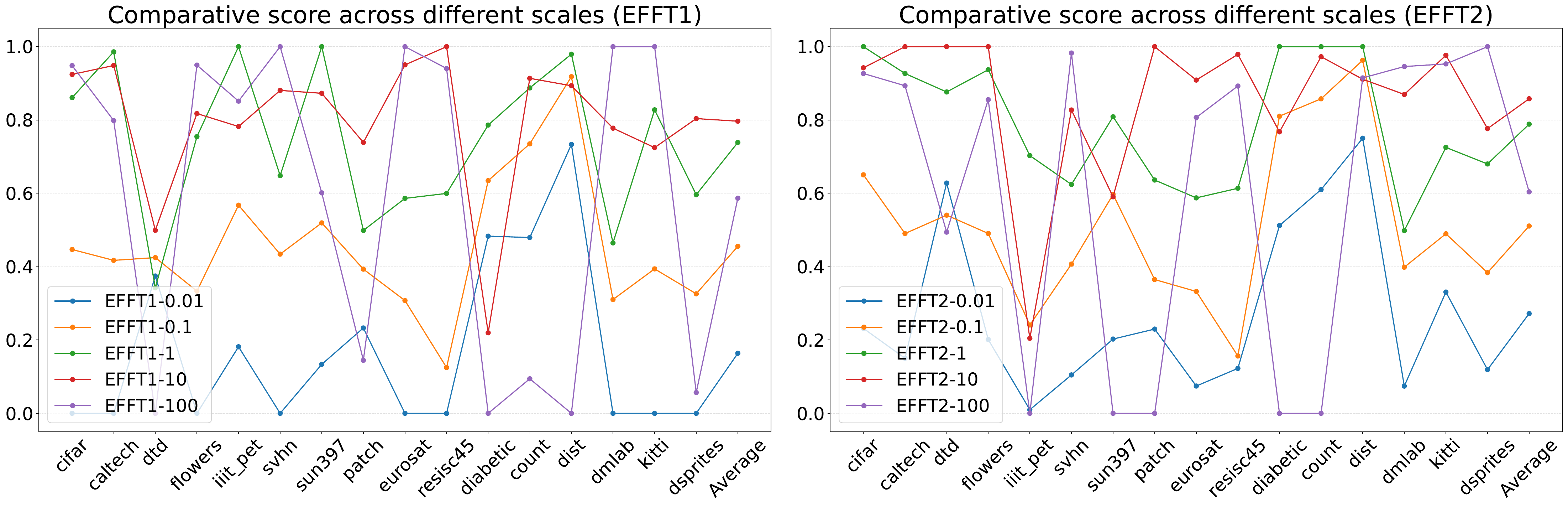}
	\caption{We summarize the performance of two EFFT methods across 16 datasets in VTAB-1K. For clarity, we normalize the best method-scale pair to 1 and the worst to 0, highlighting differences between methods. These methods are presented in two sub-graphs based on the $EFFT$ type, emphasizing variations within each method.}
	\label{fig:fig7}
\end{figure*}

\paragraph{Results.}
$EFFT$ exhibits significant improvements across all three categories of datasets, achieving SOTA results with a categorical average score of 75.9\% and an overall average score of 73.6\%. In 19 tasks, our $EFFT^1_{\leq 32}$ and $EFFT^2_{\leq 32}$ outshine competing methods, securing top performance in 14 tasks while utilizing only 0.16M and 0.24M trainable parameters, respectively. Notably, the latter configuration substantially surpasses existing mainstream approaches. 

Employing a scale-wise hyper-parameter search strategy, we managed to reduce the search space by 52\% compared to the traditional Factor-Tuning method. Furthermore, we eliminate the need for hyper-architecture search in $NOAH$ to identify optimal hyper-parameters, substantially conserving computational resources and enhancing training efficiency.

\begin{figure} 
	\includegraphics[width=\linewidth]{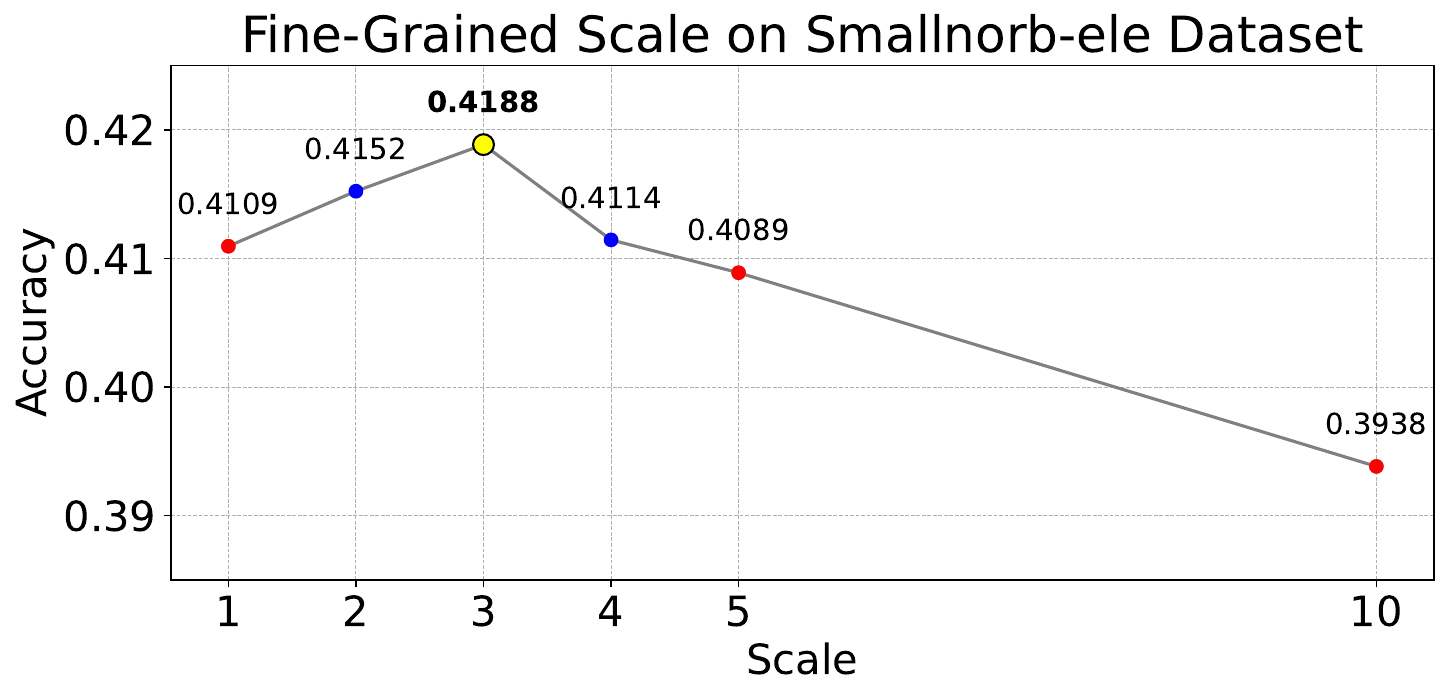}
	\caption{A figure illustrates the significance of fine-grained scales in determining optimal hyper-parameters. Given $EFFT$'s sensitivity to scale, a more optimal scale likely exists between our broadly selected values of s.}
	\label{fig:fig8}
\end{figure}

Our further experiments indicate that $EFFT$, along with other tensorization-decomposition methods, exhibit sensitivity to scale and rank. As illustrated in Figures \ref{fig:fig7} and \ref{fig:fig8}, we make experiments with a more fine-grained scale which concludes that the optimal scale hides between our roughly swept scales. This suggests that by allocating more computational resources to identify optimal hyper-parameters or by making them trainable, we can enhance the model's potency. 

\begin{table}[]
\begin{tabular}{C{1.3cm}|C{1.2cm}C{1cm}C{1cm}C{1cm}}
\toprule[1.5pt]
Name   & Average       & Nat.        & Spe.   & Str.    \\ \midrule[1.5pt]
Full   & 68.9          & 68.2          & 83.5          & 55.0            \\
Linear & 64.8          & 75.8          & 84.1          & 34.4          \\
$LoRA$   & 76.5          & 83.0            & 86.4          & 60.2          \\
$SSF$    & 74.6          & 81.9          & 86.5          & 55.3          \\
$FacT_{TK}$   & \textbf{77.4} & 83.1          & 86.9          & \textbf{62.1} \\
$EFFT_2$   & 77.0          & \textbf{83.3} & \textbf{87.2} & 60.4          \\ \bottomrule[1.5pt]
\end{tabular}
\caption{A comparative analysis of various PEFT techniques within the Swin Transformer architecture reveals that, despite a comparable number of trainable parameters, the $EFFT_2$ also yields convincing results.}
\label{tab:table2}
\end{table}

\begin{table*}[]
\begin{tabular}{C{1.25cm}|C{1cm}C{0.54cm}C{1cm}C{0.54cm}C{1cm}C{0.54cm}C{1cm}C{0.6cm}|C{1cm}C{0.54cm}C{1cm}C{0.54cm}}
\toprule[1.5pt]
\multirow{2}{*}{Name} & \multicolumn{2}{c}{Swin Tiny}            & \multicolumn{2}{c}{Swin Small}           & \multicolumn{2}{c}{Swin Base}            & \multicolumn{2}{c|}{Swin Large}          & \multicolumn{2}{c}{ViT Large}            & \multicolumn{2}{c}{ViT Huge}             \\
                  & \#Param & Ave.                           & \#Param & Ave.                           & \#Param & Ave.                           & \#Param & Ave.                           & \#Param & Ave.                           & \#Param & Ave.                           \\ \midrule[1.5pt]
Full              & 27.6M  & 65.1                           & 48.8M  & 66.6                           & 86.9M  & 68.9                           & 1.95B   & 71.3                           & 0.30B   & 60.9                           & 0.63B   & 68.3                           \\
Linear            & 0       & 64.2                           & 0       & 64.2                           & 0       & 64.8                           & 0       & 66.0                           & 0       & 60.2                           & 0       & 59.0                           \\ \midrule[1.5pt]
$LoRA$              & 0.26M   & 71.3                           & 0.36M   & 75.6                           & 0.43M   & 76.5                           & 1.47M   & 77.3                           & 0.39M   & \textbf{75.8} & 0.65M   & 72.1                           \\
$EFFT_2$              & 0.28M   & \textbf{72.1} & 0.37M   & \textbf{76.0} & 0.49M   & \textbf{77.0} & 1.51M   & \textbf{77.8} & 0.36M   & \textbf{75.8} & 0.59M   & \textbf{73.3} \\ \bottomrule[1.5pt]
\end{tabular}
\caption{We conducted experiments across a range of scales within each Vision Transformer variant. The results demonstrate that $EFFT$ consistently outperforms $LoRA$, particularly within hierarchical structures.}
\label{tab:table3}
\end{table*}

From Figure \ref{fig:fig8}, scales of 10 and 1 consistently perform well, often achieving the best results. In contrast, a scale of 100 exhibits variable performance, ranking either best or worst. The remaining scales show average or low performance. Given the consistently high performance of scale 10 across datasets, with a typical score of 0.8, we recommend it for further transfer learning due to its proven robustness and transferability.

\subsection{Scaling Up with Vision Transformer Variants}
We evaluated our method's efficiency by scaling up Vision Transformers (ViTs) and Hierarchical Transformers. For the conventional scaling of ViTs, we selected the ViT-L/16 and ViT-H/14 models as benchmarks. In the case of Hierarchical Transformers, we included Swin-T, Swin-S, Swin-B, and Swin-L in our analysis. Our methodology adhered to the previously established fine-tuning protocols. Notably, as the Swin Transformer's hidden dimension increases with depth, each PEFT method emulates this pattern by enlarging the dimensionality by a factor of $2\times$ for each successive layer.

\begin{table}[]
\begin{tabular}{c|cc|cc}
\toprule[1.5pt]
\multirow{2}{*}{Layers} & \multicolumn{2}{c|}{$EFFT_1$} & \multicolumn{2}{c}{$EFFT_2$} \\
                        & MHSA         & FFN         & MHSA        & FFN         \\ \midrule[1.5pt]
All                     & -1.54\%      & -1.85\%     & -1.69\%     & -2.32\%     \\
0-2                     & 0.74\%       & -2.51\%     & -0.72\%     & -1.79\%     \\
2-4                     & -0.87\%      & -0.82\%     & -0.56\%     & 0.12\%      \\
5-8                     & -1.32\%      & -3.52\%     & -1.11\%     & -1.24\%     \\
9-11                    & -0.73\%      & -1.19\%     & 0.91\%      & -1.48\%     \\ \bottomrule[1.5pt]
\end{tabular}
\caption{We examined the influence of MHSA and FFN on the overall model and specific layers using the $EFFT_1$ and $EFFT_2$ methods. Interestingly, in certain layers, either MHSA or FFN alone outperformed the combined fine-tuning of both, indicating greater effectiveness.}
\label{tab:table4}
\end{table}

\begin{figure} 
	\includegraphics[width=\linewidth]{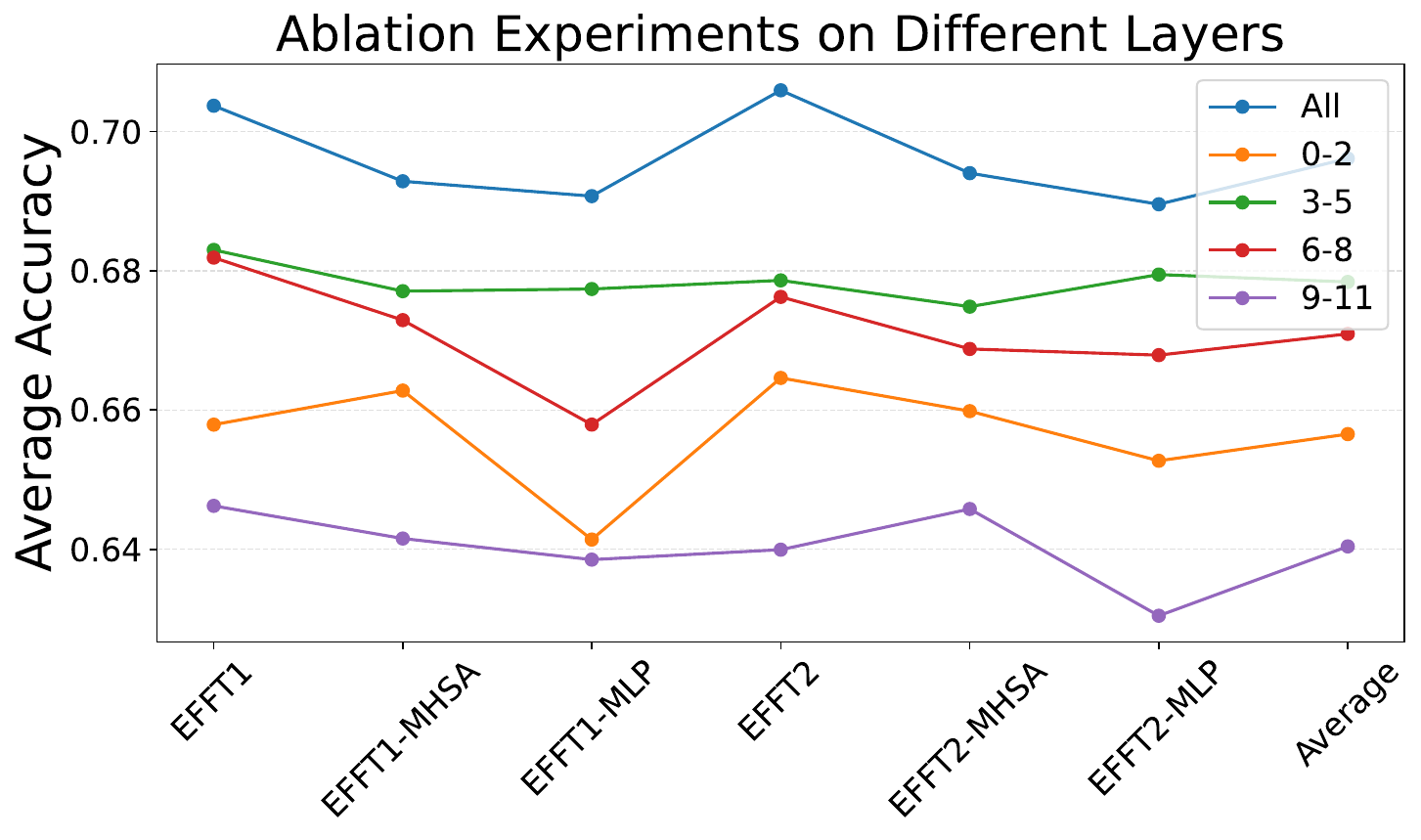}
	\caption{A review of analyzing layer impact on model performance.  Notably, the middle layers appear pivotal for fine-tuning, whereas the shallowest and deepest layers contribute minimally. This observation contrasts with findings from AdaLoRA.}
	\label{fig:fig9}
\end{figure}

\begin{table}[]
\centering
\begin{tabular}{c|cc}
\toprule[1.5pt]
Name           & ViT                 & Swin                \\ \midrule[1.5pt]
Linear & 57.7                & 68.9          \\
Full            & 69                  & 64.7          \\
$r_1 \neq r_2$      & 73.8          & 73.9          \\
$r_1 = r_2$       & \textbf{74.2} & \textbf{75.3} \\ \bottomrule[1.5pt]
\end{tabular}
\caption{Our ablation study indicates that setting $r_1 = r_2$ yields superior performance compared to $r_1 \neq r_2$  in both ViT and Swin Transformer.}
\label{tab:table5}
\end{table}

As depicted in Tables \ref{tab:table2} and \ref{tab:table3}, $EFFT$ outperforms $LoRA$ across both smaller and larger Vision Transformer variants, with a more pronounced advantage observed in the hierarchical Swin Transformer models. This can be attributed to the significant variation in intrinsic features across the layers of the Swin Transformer, necessitating the use of distinct $EFFT$ configurations.

\begin{figure*}[htbp]
        \centering
        \hspace*{-0.6cm}
	\includegraphics[width=1.08\linewidth]{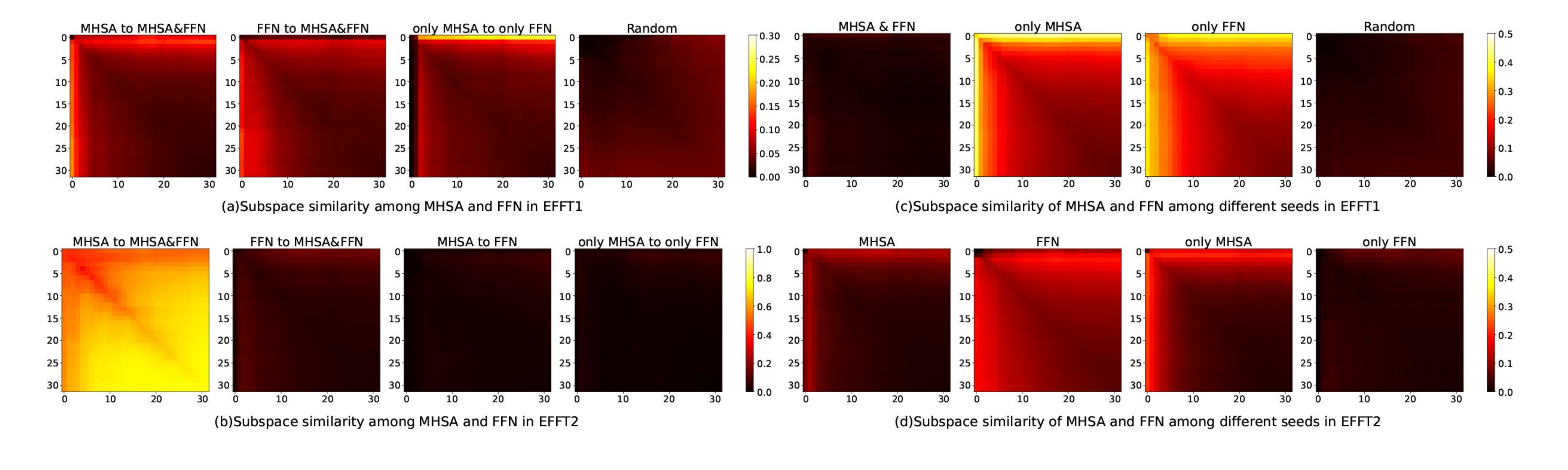}
	\caption{We conducted a thorough analysis of subspace similarities across different seeds and blocks. To mitigate the influence of similarities inherent in random Gaussian matrices, we adjusted for this random similarity before calculating the subspace similarities.}
	\label{fig:fig10}
\end{figure*}

\subsection{Ablation Studies}

We aim to identify the optimal layers within the ViT for EFFT adaptation and whether MHSA or FFN blocks contribute more to our SOTA performance. To this end, we conduct ablation studies covering all categories in VTAB-1K. Throughout these studies, we consistently employ ranks and scales that have previously exhibited optimal performance. We further categorize the 12 layers into 4 distinct groups, focusing our training on either MHSA or FFN to discern which blocks within specific layers contribute most effectively, as illustrated in Figure \ref{fig:fig9} and Table \ref{tab:table4}.

Our experimental results indicate that the MHSA block consistently outperforms FFN across nearly all methods and layers. Intriguingly, in certain scenarios, focusing solely on MHSA yields better results than training the entire model. The underlying cause of this observation remains an open question and further leads us to dive into the analysis of subspace similarities.

We also do ablation experiments on the size of the additional tensor for FFN, where we set $r_2 = 4r_1$. As we illustrated in Table \ref{tab:table5}, unequally reserving space for decomposition does not make any improvement as we experiment on ViT and Swin Transformer.

\subsection{Subspace Similarity}
To delve deeper into why fine-tuning MHSA is better than FFN, we examined subspace similarities between MHSA and FFN across various layer groups. To ensure the robustness of our findings, we incorporated two random Gaussian matrices, ensuring the detected similarities weren't coincidental.

Figure \ref{fig:fig10} illustrates a notable interconnection between MHSA and FFN. Their subspace similarities exceed 30\% in subspaces, underscoring the efficacy of $EFFT_1$ in achieving top-tier results and proving the feasibility of training MHSA and FFN together. In the subspace of $EFFT_2$, a striking similarity is evident between MHSA-only and comprehensive fine-tuning. These findings suggest that MHSA-only fine-tuning closely mirrors the subspace of comprehensive fine-tuning, reinforcing MHSA's superiority over FFN and answering the questions in our ablation studies.

In experiments with varied seeds, the $EFFT_1$ subspace for comprehensive fine-tuning displayed marked differences, while MHSA-only and FFN-only exhibited consistent subspace similarities. This suggests that comprehensive fine-tuning in $EFFT_1$ hasn't reached convergence for optimal. Conversely, $EFFT_2$, which trains MHSA and FFN separately, achieves convergence easily with higher subspace similarity. The comparison between comprehensive and FFN-only fine-tuning further indicates that $EFFT_2$'s comprehensive fine-tuning aids FFN parameter convergence. These observations partly elucidate why $EFFT_2$ consistently outperforms $EFFT_1$ across various transfer learning datasets, achieving optimal convergence in fewer steps.

\section{Conclusions and Limitations}

We propose $EFFT$, an effective plugin designed to fine-tune Vision Transformers, specifically addressing the challenges associated with inner- and cross-layer redundancy within MHSA and FFN. Our evaluations on VTAB-1K datasets demonstrate that $EFFT$ outclasses existing PEFT methods. Additionally, we conduct ablation studies to identify the tensor components that most significantly influence performance on specific downstream tasks. We further explore the subspace similarities of the additional tensor, contributing to a more comprehensive understanding of tensorization-decomposition methods.

This study also underscores several promising avenues for future exploration. Delving deeper into inner- and cross-layer redundancy could pave the way for the development of more efficient, redundancy-aware architectures.

\section*{Acknowledgement}
We would like to express our sincere appreciation to Tong Yao from Peking University (PKU) and Professor Yao Wan from Huazhong University of Science and Technology (HUST) for their invaluable contributions and guidance in this research.

\bibliography{anthology,custom}
\bibliographystyle{acl_natbib}




\end{document}